\documentclass[10pt,twocolumn]{article}

\usepackage[a4paper,margin=2cm,columnsep=0.6cm]{geometry}
\usepackage{titlesec}
\usepackage{authblk}

\usepackage[T1]{fontenc}
\usepackage[utf8]{inputenc}
\usepackage{lmodern}
\usepackage{microtype}

\usepackage{amsmath,amssymb,amsfonts}

\usepackage{booktabs}
\usepackage{multirow}
\usepackage{graphicx}
\usepackage{float}
\usepackage{caption}
\usepackage{subcaption}
\usepackage{array}

\usepackage{tikz}
\usetikzlibrary{arrows.meta,positioning,calc,fit,backgrounds}

\usepackage{algorithm}
\usepackage{algorithmic}

\usepackage{listings}
\usepackage[hidelinks,colorlinks=true,citecolor=blue!60!black,linkcolor=blue!60!black,urlcolor=blue!60!black]{hyperref}
\usepackage[numbers,sort&compress]{natbib}

\hypersetup{
  pdftitle={Micro-Diffusion Compression: Binary Tree Tweedie Denoising for Online Probability Estimation},
  pdfauthor={Roberto Tacconelli},
  pdfsubject={Lossless Data Compression},
  pdfkeywords={lossless compression, diffusion, Tweedie denoising, binary tree, PPM, arithmetic coding, context modeling},
  pdfusetitle=false,
}

\usepackage{xcolor}
\usepackage{enumitem}
\setlist{nosep,leftmargin=*}

\titleformat{\section}{\large\bfseries}{\thesection.}{0.5em}{}
\titleformat{\subsection}{\normalsize\bfseries}{\thesubsection}{0.5em}{}
\titleformat{\subsubsection}{\normalsize\itshape}{\thesubsubsection}{0.5em}{}

\newcommand{\mdc}{\textsc{Midicoth}}
\newcommand{\bpb}{\text{bpb}}

\title{\textbf{Micro-Diffusion Compression:\\Binary Tree Tweedie Denoising for Online Probability Estimation}}

\author[1]{Roberto Tacconelli}
\affil[1]{Independent Researcher\\
\texttt{tacconelli.rob@gmail.com}}

\date{}

\begin{document}
\maketitle
\thispagestyle{empty}

% =====================================================================
% ABSTRACT
% =====================================================================
\begin{abstract}
We present \mdc{}\footnote{The name \textsc{Midicoth} derives from
\emph{mi}cro-\emph{di}ffusion \emph{co}mpression \emph{th}reshold.},
a lossless compression system that introduces
\emph{micro-diffusion}---a multi-step score-based reverse diffusion
process implementing Tweedie's empirical Bayes formula---into a
cascaded statistical modeling pipeline. The system treats a context
model's Jeffreys-prior smoothing as a shrinkage operator that pulls the
empirical distribution toward uniform, with effective noise level
$\gamma = 128/(C{+}128)$ and reverses it through a
binary tree decomposition: each 256-way byte prediction is decomposed
into eight binary decisions (MSB to LSB), and at each tree node the
additive Tweedie correction
$\delta = E[\theta|\hat{p}] - E[\hat{p}]$ is estimated
nonparametrically and applied across $K=3$ successive denoising steps
with independent score tables.

The \mdc{} pipeline cascades five fully online, parameter-free layers:
(1)~an order-0 through order-4 adaptive PPM model with PPMC-style
exclusion and Jeffreys prior;
(2)~an extended match model for long-range repetition;
(3)~a trie-based word model with bigram prediction;
(4)~a high-order context model aggregating orders~5--8; and
(5)~the micro-diffusion layer with binary tree Tweedie denoising,
applied as the final post-blend correction step.

On the enwik8 benchmark (100\,MB Wikipedia), \mdc{} achieves
\textbf{1.753\,\bpb{}} (21.9\,MB), outperforming xz~-9
(1.989\,\bpb{}) by \textbf{11.9\%} and all evaluated dictionary-based
compressors, without any pre-trained neural network, training data, or
GPU. On alice29.txt (152\,KB Canterbury Corpus), \mdc{} achieves
\textbf{2.119\,\bpb{}}, outperforming xz~-9 (2.551\,\bpb{}) by 16.9\%.

An ablation study demonstrates that each component contributes
measurably: PPMC exclusion provides a strong base, the match model
contributes up to 5.5\% on repetitive data, and the post-blend
micro-diffusion layer---applied after all model blending---adds
2.3--2.7\% by correcting systematic biases in the fully blended
distribution. The binary tree decomposition enables data-efficient
calibration (each node produces a binary outcome), while multi-step
denoising enables residual correction across steps.
The entire system is
implemented in $\sim$2{,}000 lines of C with no external dependencies,
compresses at $\sim$60\,KB/s on a single CPU core, and is fully
deterministic with bit-exact encoder--decoder symmetry.
\end{abstract}

\medskip
\noindent\textbf{Keywords:} lossless compression, diffusion processes,
Tweedie denoising, binary tree, PPM, arithmetic coding, context modeling

% =====================================================================
\section{Introduction}
\label{sec:intro}
% =====================================================================

Lossless data compression reduces storage and transmission costs by
exploiting statistical redundancy in the source. The quality of
compression is bounded by how well the compressor can predict the next
symbol: a model that assigns probability $p$ to the correct symbol
emits $-\log_2 p$ bits, so sharper predictions yield shorter codes.
This prediction--compression duality, implicit in information
theory~\citep{shannon1948} and made explicit by arithmetic
coding~\citep{witten1987}, has motivated progressively richer models
---from fixed Huffman tables~\citep{huffman1952}, through
dictionary-based methods (LZ77~\citep{ziv1977}, LZMA/xz), to the
adaptive context models of PPM~\citep{cleary1984} and the massive
context-mixing ensembles of PAQ/CMIX~\citep{mahoney2005, knoll2012}.

Despite decades of progress, a fundamental bottleneck persists in
adaptive statistical compressors. Models such as PPM estimate the
next-symbol distribution by counting byte occurrences within matching
contexts, smoothed by a prior (e.g., Jeffreys' 0.5 per
symbol~\citep{jeffreys1946}). When few observations are available,
the prior dominates: the predicted distribution is far flatter than
the true source distribution, and bits are wasted encoding symbols
that the model could, in principle, predict more sharply. When many
observations are available, the prior becomes negligible and the
distribution is already well-calibrated---but the model has no
mechanism to distinguish these two regimes. The distribution is
simply the normalized count vector, regardless of whether it is
prior-dominated or data-dominated.

We observe that this bottleneck can be framed as a
\emph{denoising problem}: Jeffreys smoothing acts as a shrinkage operator that pulls the
empirical distribution toward uniform---and this shrinkage can be
reversed using Tweedie's empirical Bayes formula~\citep{efron2011}.
This motivates \emph{micro-diffusion}---a post-prediction correction
layer that decomposes each 256-way prediction into a binary tree of
eight decisions (MSB to LSB), estimates the Tweedie score function
nonparametrically at each node via calibration tables, and applies
the additive correction $\hat{p}' = \hat{p} + \delta$ across $K=3$
successive denoising steps with independent score tables.

We make the following contributions:

\begin{enumerate}
\item \textbf{Binary tree Tweedie denoising.} A multi-step score-based
reverse diffusion process that decomposes byte-level predictions into
binary decisions and estimates the additive Tweedie correction
$\hat{\delta} \approx \sigma^2 \cdot s(\hat{p})$
at each tree node via nonparametric score tables ($\sim$4.7\,MB,
155{,}520 entries) indexed by step,
bit context, model order, distribution shape, and noise level
(Section~\ref{sec:diffusion}).

\item \textbf{Binary tree decomposition for calibration.} A coarse-to-fine
hierarchy (MSB to LSB) that converts 256-way calibration into binary
calibration, dramatically improving data efficiency and enabling
enriched context modeling through parent bit values
(Section~\ref{sec:bintree}).

\item \textbf{Five-layer cascaded pipeline.} A carefully ordered
pipeline combining PPM, match, word, high-order context models, and
Tweedie denoising as the final post-blend correction step. Each
layer operates on the output of the previous one; crucially, placing
Tweedie \emph{after} all model blending allows it to correct
systematic biases introduced by the full ensemble, not just the base
PPM (Section~\ref{sec:pipeline}).

\item \textbf{Comprehensive ablation.} A controlled study
demonstrating each component's marginal contribution across multiple
file sizes and types, with cross-implementation validation
(Section~\ref{sec:ablation}).
\end{enumerate}

Together, these contributions yield a pure-statistical compressor
that outperforms xz~-9 on \emph{all} tested inputs (\textbf{1.753 vs.\
1.989\,\bpb{}} on enwik8, \textbf{2.119 vs.\ 2.551\,\bpb{}} on
alice29.txt) without any neural network, training data,
or GPU---narrowing the gap to heavyweight context-mixing systems
like PAQ and CMIX.

% =====================================================================
\section{Related Work}
\label{sec:related}
% =====================================================================

\subsection{Prediction by Partial Matching (PPM)}

PPM~\citep{cleary1984} maintains adaptive context models of varying
orders (typically 0--5) and predicts the next symbol by falling back
from the highest matching order to lower ones. PPMD+ achieves
$\sim$2.0\,\bpb{} on English text. PPM's fundamental limitation is
that it treats each context's count array as a direct probability
estimate, with no mechanism to sharpen confident predictions or
correct systematic biases---the distribution is the normalized count
vector, nothing more.

\subsection{Context Mixing (PAQ/CMIX)}

The PAQ family~\citep{mahoney2005} extends PPM with context mixing:
blending predictions from hundreds of specialized models via neural
networks at the bit level. PAQ8px achieves $\sim$1.27\,\bpb{} on
enwik8~\citep{mahoney2023ltcb}. CMIX~\citep{cmix2024} incorporates
LSTM networks alongside 2{,}000+ context models, reaching
$\sim$1.17\,\bpb{} on enwik8---but at extreme cost (16--64\,GB RAM,
hours of compression time). These systems demonstrate that
post-prediction correction (mixing, SSE) can dramatically improve
compression, motivating our work on a lightweight alternative.

\subsection{Secondary Symbol Estimation}

SSE, introduced in the PAQ family, applies a piecewise-linear
correction to predicted probabilities via Adaptive Probability Maps
(APMs). Each APM maintains a lookup table mapping input logit to
corrected probability, updated online via gradient descent. SSE
is the intellectual precursor to micro-diffusion: both are
post-prediction correction layers. We evaluated a two-stage SSE chain
and found that it interferes with Tweedie denoising---both learn
probability recalibration, and their combination produces worse
results than Tweedie alone. The key difference is that SSE uses
gradient-based piecewise-linear maps, while our approach directly
implements Tweedie's formula via additive corrections estimated from
sufficient statistics (hit rate and average prediction).

\subsection{Temperature Scaling and Tweedie's Formula}

Temperature scaling is widely used in deep learning for calibrating
neural network predictions~\citep{guo2017}. Tweedie's
formula~\citep{efron2011} provides the optimal Bayes estimator for
exponential family models under squared-error loss:
$\hat{\theta} = y + \sigma^2 \nabla \log m(y)$, where $m(y)$ is the
marginal density. Our work applies this principle by analogy: Jeffreys smoothing can be interpreted as a shrinkage operator that
pulls the empirical distribution toward uniform, with effective
noise level $\gamma = 128/(C{+}128)$ playing the role of $\sigma^2$.
Since this shrinkage is a convex mixture toward uniform
rather than additive Gaussian noise, the Tweedie link is conceptual
rather than exact. Nevertheless, calibration tables estimate the
additive correction $\hat{\delta} = \widehat{E}[\theta|\hat{p}]
- \widehat{E}[\hat{p}]$ nonparametrically, which approximates the
Tweedie term $\sigma^2 \cdot s(\hat{p})$.

\subsection{Diffusion Models}

Diffusion models~\citep{ho2020, song2021} learn to reverse a gradual
noising process by estimating the score function $\nabla \log p_t(x)$
at each noise level~$t$. Our micro-diffusion layer instantiates
this framework: Jeffreys smoothing acts as a shrinkage operator toward uniform
whose effective noise level $\gamma = 128/(C{+}128)$ depends on the
sample size; we treat this shrinkage as an equivalent forward
noise process, and the calibration tables estimate the
score function $s(\hat{p}, \gamma)$ nonparametrically---conditioned
on the noise level via the confidence dimension. The additive
Tweedie correction $\hat{p}' = \hat{p} + \delta$ is the score-based
denoising update, and the $K=3$ steps constitute multi-step reverse
diffusion with independent score estimators at each step.

\subsection{LLM-Based Compression}

Del\'etang et al.~\citep{deletang2024} formalized the LLM-as-compressor
paradigm. Practical implementations include FineZip~\citep{huang2024finezip},
ts\_zip~\citep{bellard2023}, and Nacrith~\citep{tacconelli2025}.
These achieve 0.9--1.1\,\bpb{} on enwik8 using pre-trained transformer
models. \mdc{} operates in a different regime: no neural network, no
training data, no GPU---achieving 1.753\,\bpb{} with purely adaptive
statistical models running on a single CPU core.

% =====================================================================
\section{Method}
\label{sec:method}
% =====================================================================

\subsection{Pipeline Overview}
\label{sec:pipeline}

\mdc{} processes input one byte at a time. For each byte position~$i$,
five layers produce a probability distribution $P(s)$ over the 256
possible byte values, which is then fed to an arithmetic coder.
The layers execute in a fixed cascade (Figure~\ref{fig:pipeline}):

\begin{figure}[t]
\centering
\begin{tikzpicture}[
  node distance=0.32cm,
  box/.style={draw, rounded corners=2pt, minimum width=3.0cm,
              minimum height=0.52cm, font=\scriptsize\sffamily,
              fill=#1, text=white, align=center},
  box/.default=black!65,
  arr/.style={-{Stealth[length=4pt]}, thick, black!70},
  lbl/.style={font=\tiny\sffamily, text=black!55},
]

% Main vertical chain
\node[box=black!50] (input) {Input byte $x_i$};

\node[box=blue!60!black, below=of input] (ppm)
  {PPM (orders 0--4)};

\node[box=black!45, below=0.22cm of ppm,
      minimum height=0.38cm] (norm1) {\tiny Normalize};

\node[box=teal!70!black, below=of norm1] (match)
  {Match blend};

\node[box=teal!70!black, below=0.22cm of match] (word)
  {Word blend};

\node[box=teal!70!black, below=0.22cm of word] (hctx)
  {HighCtx blend};

\node[box=red!65!black, below=of hctx,
      minimum width=3.0cm, minimum height=0.65cm] (tweedie)
  {Micro-Diffusion\\[-1pt]{\tiny Tweedie $\times\,3$ steps}};

\node[box=black!45, below=0.22cm of tweedie,
      minimum height=0.38cm] (norm2) {\tiny Normalize};

\node[box=black!70, below=of norm2] (arith)
  {Arithmetic Coder};

\node[box=black!50, below=of arith] (output)
  {Compressed bits};

% Main spine arrows
\draw[arr] (input) -- (ppm);
\draw[arr] (ppm)   -- node[lbl, right, xshift=1pt]
  {$\mathbf{p},\;C,\;k$} (norm1);
\draw[arr] (norm1)  -- (match);
\draw[arr] (match)  -- (word);
\draw[arr] (word)   -- (hctx);
\draw[arr] (hctx)   -- (tweedie);
\draw[arr] (tweedie) -- (norm2);
\draw[arr] (norm2)  -- (arith);
\draw[arr] (arith)  -- (output);

% Update feedback (left side, away from everything)
\draw[arr, dashed, black!40]
  ([xshift=-4pt]arith.west) -- ++(-0.5,0)
  |- node[lbl, left, pos=0.20] {update $x_i$}
  ([xshift=-4pt]ppm.west);

% Bracket: blending group
\begin{scope}[on background layer]
  \node[draw=teal!30, fill=teal!4, rounded corners=3pt,
        fit=(match)(word)(hctx),
        inner xsep=5pt, inner ysep=3pt] (blendgrp) {};
\end{scope}
\node[lbl, right=0.1cm of blendgrp] {blending};

\end{tikzpicture}
\caption{The \mdc{} compression pipeline. Each byte passes through
PPM prediction, three sequential blending layers (Match, Word, HighCtx),
and the micro-diffusion layer (3-step Tweedie denoising with
James-Stein shrinkage) before arithmetic coding. All models are
updated after encoding (dashed arrow).}
\label{fig:pipeline}
\end{figure}
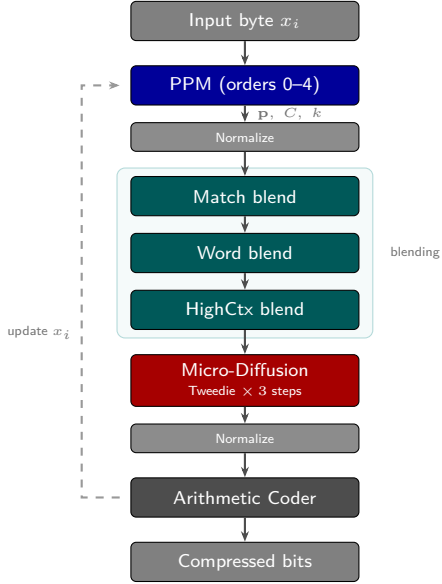

\begin{algorithm}[t]
\caption{\mdc{} compression pipeline (per byte)}
\label{alg:pipeline}
\begin{algorithmic}[1]
\REQUIRE byte stream $(x_1, \ldots, x_n)$
\STATE Initialize: PPM, Match, Word, HighCtx, Tweedie, ArithEncoder
\FOR{$i = 1$ to $n$}
  \STATE $\mathbf{p}, C, k \leftarrow \text{PPM.predict}()$ \COMMENT{distribution, confidence, order}
  \STATE $\mathbf{p} \leftarrow \text{Normalize}(\mathbf{p})$
  \STATE $\mathbf{p} \leftarrow \text{Match.blend}(\mathbf{p})$ \COMMENT{long-range repetition}
  \STATE $\mathbf{p} \leftarrow \text{Word.blend}(\mathbf{p})$ \COMMENT{word completion}
  \STATE $\mathbf{p} \leftarrow \text{HighCtx.blend}(\mathbf{p})$ \COMMENT{orders 5--8}
  \STATE $\mathbf{p} \leftarrow \text{Tweedie.denoise}(\mathbf{p}, C, k)$ \COMMENT{post-blend denoising}
  \STATE $\mathbf{p} \leftarrow \text{Normalize}(\mathbf{p})$
  \STATE $\text{cumfreqs} \leftarrow \text{probs\_to\_cumfreqs}(\mathbf{p})$
  \STATE ArithEncoder.encode(cumfreqs, $x_i$)
  \STATE Update all models with observed byte $x_i$
\ENDFOR
\RETURN ArithEncoder.finish()
\end{algorithmic}
\end{algorithm}

Decompression executes an identical pipeline; the arithmetic decoder
recovers each symbol from the same probability distribution, ensuring
perfect lossless reconstruction. All layers are fully online and
adaptive---no pre-training or offline parameter estimation is required.

\subsection{Base Model: Adaptive PPM with Jeffreys Prior}
\label{sec:ppm}

The base model maintains order-0 through order-4 adaptive context
models. Each order uses an open-addressing hash table mapping
FNV-1a~\citep{fowler1991} context hashes to 256-element count arrays
stored in double precision.

\paragraph{Prediction with PPMC Exclusion.}
The model uses a fallback chain from highest to lowest order with
PPMC-style exclusion~\citep{moffat1990ppm}. When a higher-order
context escapes to a lower order, symbols already assigned probability
at the higher order are \emph{excluded} from the lower-order
distribution:

\begin{equation}
P(s) = \sum_{k} \left(\prod_{j>k} p_{\text{esc},j}\right)
  (1 - p_{\text{esc},k}) \cdot \frac{c_k(s)}{n_k}
  \cdot \mathbf{1}[s \notin \mathcal{E}_k]
\label{eq:ppm_predict}
\end{equation}

where $n_k$ is the total real observation count (counts minus prior)
for non-excluded symbols at order~$k$, $d_k$ is the number of distinct
observed symbols, and the escape probability follows Method~C:
$p_{\text{esc},k} = d_k / (n_k + d_k)$.
The exclusion set $\mathcal{E}_k$ accumulates all symbols seen at
orders above~$k$. Remaining probability mass after all orders is
distributed uniformly among symbols not yet excluded.

\paragraph{Prior.}
New contexts are initialized with the Jeffreys prior~\citep{jeffreys1946}:
$c_k(s) = 0.5$ for all $s$, giving an initial total of
$T_0 = 256 \times 0.5 = 128$. This is the maximum-entropy
non-informative prior for multinomial estimation, balancing between
the uniform (Laplace) prior and the maximum-likelihood estimate. The
Jeffreys prior's strength---128 pseudo-counts across 256 symbols---is
central to understanding micro-diffusion's benefit: on low-count
contexts, the prior dominates and the distribution is flatter than
the true source distribution. The Jeffreys prior is chosen for
its invariance under reparameterization and its guarantee of
non-zero probabilities for all symbols---the standard smoothing
choice in PPM implementations~\citep{cleary1984}.

\paragraph{Update.}
After each observed byte, counts are incremented for all available
orders (0 through~4). Hash tables auto-grow at 60\% load factor using
open addressing with linear probing. The confidence value returned
to the Tweedie denoising layer is the total count $C = \sum_s c_k(s)$
of the matched context.

\subsection{Micro-Diffusion Layer: Binary Tree Tweedie Denoising}
\label{sec:diffusion}

\subsubsection{Motivation: Ensemble Calibration}

The micro-diffusion layer operates on the \emph{fully blended}
distribution (after PPM, match, word, and high-order context models
have all contributed). While the original motivation arises from
PPM's prior-dilution problem, placing Tweedie after all blending
allows it to correct systematic biases introduced by the entire
ensemble---not just PPM's Jeffreys prior.

Consider a PPM context that has been observed $n$ times, with the true
next-symbol distribution $\mathbf{q}$. With Jeffreys prior, the
model's estimate is:

\begin{equation}
\hat{p}(s) = \frac{n \cdot q(s) + 0.5}{n + 128}
\label{eq:prior_dilution}
\end{equation}

This is a convex mixture between the true distribution and the uniform:
$\hat{\mathbf{p}} = \lambda \mathbf{q} + (1-\lambda) \mathbf{u}$,
where $\lambda = n/(n+128)$. When $n$ is small (e.g., $n = 5$),
$\lambda = 5/133 \approx 0.04$, so the prior contributes 96\% of the
total mass. To illustrate concretely at the binary tree level:
consider the root node (level~0), where the true probability of
going right is $q_R = 0.9$. Each subtree contains 128~symbols with
Jeffreys prior $0.5$ each, contributing 64 pseudo-counts per side.
The smoothed estimate is
$\hat{p}_R = (5 \cdot 0.9 + 64) / (5 + 128)
= 68.5/133 \approx 0.515$,
pulled almost to 0.5---costing $-\log_2(0.515) = 0.96$ bits instead
of $-\log_2(0.9) = 0.15$ bits, a $6.3\times$ overhead at this single
node. Across eight tree levels, such dilution compounds severely.

We frame this as a denoising problem: Jeffreys smoothing is a
shrinkage operator that pulls the empirical distribution toward
uniform, and our goal is to reverse this shrinkage.

\subsubsection{Theoretical Framework: Tweedie Empirical Bayes}

Tweedie's formula~\citep{efron2011} gives the optimal Bayes estimator
for exponential family models under squared-error loss:

\begin{equation}
\hat{\theta}_i = \hat{p}_i + \sigma^2 \cdot
\frac{\partial}{\partial \hat{p}_i} \log m(\hat{\mathbf{p}})
\label{eq:tweedie}
\end{equation}

where $m(\hat{\mathbf{p}})$ is the marginal density of the observed
distribution and $\sigma^2$ is the noise variance. For exponential
family models with additive Gaussian noise, Tweedie's formula implies:

\begin{equation}
\sigma^2 \cdot s(\hat{p}) = E[\theta \mid \hat{p}] - E[\hat{p}]
\label{eq:tweedie_delta}
\end{equation}

In our setting, the shrinkage operator is a convex mixture toward
uniform ($\hat{\mathbf{p}} = \lambda \mathbf{q} + (1{-}\lambda)
\mathbf{u}$) rather than additive Gaussian noise, so
Equation~\ref{eq:tweedie_delta} holds only approximately,
with $\gamma = 128/(C{+}128)$ playing the role of~$\sigma^2$.
Nevertheless, the key insight carries over: the additive correction
$\delta = E[\theta|\hat{p}] - E[\hat{p}]$ approximates the optimal
denoising direction regardless of the noise model.

This motivates our nonparametric estimator: we partition the
observation space into bins, track the empirical hit rate
$\widehat{E}[\theta|\hat{p}]$ and average prediction
$\widehat{E}[\hat{p}]$ within each bin, and compute
$\hat{\delta} = \widehat{E}[\theta|\hat{p}] - \widehat{E}[\hat{p}]$
as a consistent estimate of the optimal additive correction.
The corrected estimate is then
$\hat{p}' = \hat{p} + \hat{\delta}$, inspired by
Equation~\ref{eq:tweedie}.

\subsubsection{Binary Tree Decomposition}
\label{sec:bintree}

Direct calibration of a 256-way distribution is data-inefficient:
each symbol has only a small fraction of the probability mass, and
calibrating 256 probabilities simultaneously requires many observations.
We instead decompose each 256-way prediction into a binary tree of
8 decisions, one per bit from MSB to LSB:

\begin{equation}
P(\text{byte} = s) = \prod_{\ell=0}^{7}
P(\text{bit}_\ell \mid \text{bits}_{0:\ell-1})
\label{eq:bintree}
\end{equation}

At each internal node of the binary tree, we compute the probability
of going right (toward higher-valued symbols) by marginalizing over
the node's subtree:

\begin{equation}
P_\ell(\text{right} \mid \text{node}) =
\frac{\sum_{s \in \text{right subtree}} P(s)}
     {\sum_{s \in \text{node's subtree}} P(s)}
\label{eq:p_right}
\end{equation}

This binary probability is then denoised by the additive Tweedie
correction (Section~\ref{sec:calib}). The binary decomposition offers two key
advantages:

\begin{enumerate}
\item \textbf{Data efficiency.} Each binary prediction has exactly two
outcomes (left/right), so calibration tables receive one observation
per prediction. With 256 symbols, byte-level calibration would need
$\sim$256$\times$ more data to achieve the same precision.

\item \textbf{Hierarchical context.} The tree structure provides a
natural coarse-to-fine hierarchy: level~0 (MSB) distinguishes broad
byte ranges (control characters vs.\ printable ASCII), while level~7
(LSB) makes fine distinctions within narrow ranges. Different levels
exhibit qualitatively different calibration patterns.
\end{enumerate}

\subsubsection{Enriched Bit Context}

To capture the distinct calibration characteristics at each tree level,
we define 27 bit contexts encoding both the level and the path taken
through higher levels:

\begin{table}[H]
\centering
\scriptsize
\setlength{\tabcolsep}{3pt}
\begin{tabular}{@{}cccl@{}}
\toprule
\textbf{Level} & \textbf{Bit} & \textbf{Ctx} & \textbf{Encoding} \\
\midrule
0 & MSB (bit 7) & 1 & No parent info \\
1 & bit 6 & 2 & Indexed by bit 7 \\
2 & bit 5 & 4 & Indexed by bits 7,6 \\
3--7 & bits 4--0 & $4{\times}5{=}20$ & Hash of higher bits $\to$ 4 groups \\
\midrule
\multicolumn{2}{@{}c}{\textbf{Total}} & \textbf{27} & \\
\bottomrule
\end{tabular}
\end{table}

This distinguishes, for example, ``LSB decision within ASCII uppercase
range'' from ``LSB decision within control characters''---different
byte ranges have very different calibration curves.

\subsubsection{Calibration Table}
\label{sec:calib}

The Tweedie correction at each binary node is determined by a
calibration table indexed by six dimensions:

\begin{equation}
\text{table}[\text{step}][\text{bctx}][\text{ord}][\text{shape}][\text{conf}][\text{pbin}]
\label{eq:calib_table}
\end{equation}

\begin{itemize}
\item \textbf{Step} $t \in \{0, 1, 2\}$: denoising iteration (Section~\ref{sec:multistep}).
\item \textbf{Bit context} $b \in \{0, \ldots, 26\}$: tree level + parent path (27 values).
\item \textbf{Order group} $o \in \{0, 1, 2\}$: PPM order $\{$-1,0,1$\}$, $\{$2,3$\}$, $\{$4+$\}$.
\item \textbf{Shape} $s \in \{0, 1, 2, 3\}$: distribution peakedness via $p_{\max}$ (thresholds 0.05, 0.15, 0.40).
\item \textbf{Confidence (noise level)} $c \in \{0, \ldots, 7\}$:
log-spaced bins of the context's observation count~$C$.
This dimension serves as \emph{noise-level conditioning}: the noise
fraction $\gamma = 128/(C{+}128)$ determines how much the prior has
diluted the distribution, so the score function
$s(\hat{p}, \gamma)$ varies with~$\gamma$---analogous to
$s_\theta(x_t, t)$ in diffusion models~\citep{ho2020}.
\item \textbf{Probability bin} $b \in \{0, \ldots, 19\}$: logit-spaced discretization of $P(\text{right})$ in $[-8, 8]$.
\end{itemize}

Total entries: $3 \times 27 \times 3 \times 4 \times 8 \times 20 = 155{,}520$,
requiring $\sim$4.7\,MB of memory (four doubles per entry).

Each entry tracks four sufficient statistics:
\begin{gather}
\text{sum\_pred}(e) = \textstyle\sum_i \hat{p}_i, \quad
\text{hits}(e) = \textstyle\sum_i \mathbf{1}[\text{went right}],
\notag \\
\text{total}(e) = N_e, \quad
\text{sum\_sq\_err}(e) = \textstyle\sum_i (b_i - \hat{p}_i)^2
\label{eq:calib_stats}
\end{gather}
where $b_i \in \{0,1\}$ is the binary outcome and $\hat{p}_i$ is the
predicted $P(\text{right})$.

The Tweedie additive correction is:
\begin{equation}
\delta = \underbrace{\frac{\text{hits}(e)}{\text{total}(e)}}_{E[\theta|\hat{p}]}
       - \underbrace{\frac{\text{sum\_pred}(e)}{\text{total}(e)}}_{E[\hat{p}]}
\label{eq:tweedie_correction}
\end{equation}

By the analogy with Equation~\ref{eq:tweedie_delta}, this $\delta$
approximates the Tweedie correction term
$\sigma^2 \cdot s(\hat{p})$, with accuracy improving as the bin
population grows.
The corrected probability is:
\begin{equation}
P'(\text{right}) = \text{clamp}\bigl(P(\text{right}) + \delta,\;
\epsilon,\; 1{-}\epsilon\bigr)
\label{eq:additive_correction}
\end{equation}

with $\epsilon = 10^{-8}$. When the model systematically
over-predicts $P(\text{right})$ ($\delta < 0$), the correction
reduces it; when under-predicting ($\delta > 0$), it increases it.
Note that the additive correction and clamping can momentarily violate
the simplex constraint (the 256 leaf probabilities may no longer sum
to~1); this is resolved by the renormalization step at the end of
each denoising iteration (Algorithm~\ref{alg:diffusion}, line~10).

\paragraph{Variance-aware James-Stein shrinkage.}
The raw Tweedie correction $\delta$ can be noisy in sparsely-populated
calibration bins, particularly early in the stream. We apply a
James-Stein-type shrinkage~\citep{james1961,efron2012} that attenuates
$\delta$ toward zero when the signal-to-noise ratio (SNR) is low:
\begin{equation}
\text{SNR} = \frac{\delta^2 \cdot N_e}{\text{Var}_e}, \quad
\delta' = \delta \cdot \min\!\Bigl(1,\; \frac{\text{SNR}}{4}\Bigr)
\label{eq:shrinkage}
\end{equation}
where $\text{Var}_e = \text{sum\_sq\_err}(e) / N_e$ is the empirical
variance of the prediction error in the bin. When $\text{SNR} \ge 4$,
the correction is applied in full; below this threshold, it is linearly
attenuated, reaching zero when the signal is indistinguishable from noise.
For bins with fewer than~10 observations, $\delta$ is set to zero entirely.

This shrinkage prevents noisy corrections from \emph{hurting}
compression---a crucial property since every miscalibrated correction
costs bits. The improvement grows with file size: on alice29 (152\,KB)
the Tweedie contribution increases from +2.46\% to +2.63\%, while on
enwik8\_3M (3\,MB) it increases from +2.71\% to +2.77\%.

\paragraph{Why additive, not multiplicative.}
The natural form of Tweedie's formula (Equation~\ref{eq:tweedie}) is
additive: $\hat{\theta} = \hat{p} + \sigma^2 \cdot s(\hat{p})$. We
verified empirically that the additive correction matches or
outperforms a multiplicative ratio correction
($P' = P \cdot r$ where $r = \text{hits}/\text{sum\_pred}$) across
all test files.

\paragraph{Prior initialization.}
Each entry is initialized with $W = 32$ pseudo-observations centered
at the bin's midpoint probability, ensuring smooth behavior before
real data arrives:
$\text{sum\_pred} = p_{\text{center}} \cdot W$,
$\text{hits} = p_{\text{center}} \cdot W$,
$\text{total} = W$,
$\text{sum\_sq\_err} = 0.25 \cdot W$ (Bernoulli variance at $p = 0.5$).

\subsubsection{Multi-Step Reverse Diffusion}
\label{sec:multistep}

A single correction pass cannot fully reverse the prior's dilution,
because the correction itself changes the distribution's shape and
probability bins. We apply $K = 3$ successive denoising steps, each
with its own independent calibration table:

\begin{algorithm}[t]
\caption{Multi-step binary tree Tweedie denoising}
\label{alg:diffusion}
\begin{algorithmic}[1]
\REQUIRE distribution $\mathbf{p} \in \Delta^{255}$, confidence $C$, PPM order $k$
\FOR{$t = 0$ to $K-1$}
  \STATE Build sum tree: $S[i] = S[2i] + S[2i{+}1]$ for $i = 255, \ldots, 1$
  \FOR{each internal node $i$ (levels 0--7)}
    \STATE $P_R \leftarrow S[2i{+}1] / S[i]$ \COMMENT{P(right) from sum tree}
    \STATE Look up calibration entry for $(t, \text{bctx}_i, k, \text{shape}, C, P_R)$
    \STATE $\delta \leftarrow \text{hits}/\text{total} - \text{sum\_pred}/\text{total}$ \COMMENT{$\approx \sigma^2 s(\hat{p})$}
    \STATE $\text{SNR} \leftarrow \delta^2 \cdot N / (\text{sum\_sq\_err}/N)$ \COMMENT{James-Stein shrinkage}
    \STATE $\delta \leftarrow \delta \cdot \min(1,\; \text{SNR}/4)$
    \STATE $P_R' \leftarrow \text{clamp}(P_R + \delta)$ \COMMENT{additive correction}
    \STATE Compute scale factors: $s_L = (1{-}P_R')/(1{-}P_R)$, $s_R = P_R'/P_R$
  \ENDFOR
  \STATE Propagate scales top-down; apply leaf scales to $\mathbf{p}$
  \STATE Renormalize $\mathbf{p}$
  \STATE Recompute shape bin from corrected $\mathbf{p}$
\ENDFOR
\RETURN $\mathbf{p}$
\end{algorithmic}
\end{algorithm}

\paragraph{Why multiple steps help.}
After step~0 applies its Tweedie correction, the distribution changes
shape (typically becoming more peaked). Step~1 observes this
partially-denoised distribution and applies a residual score
correction via its own independent calibration table. Each step
estimates the score at the \emph{current} noise level of its input,
forming a multi-step reverse diffusion chain where successive score
estimates refine the denoising. Empirically,
$K = 3$ steps reduce compressed size by 1.5--2.5\% beyond a single step.
We chose $K = 3$ because it balances correction strength against
calibration-table sparsity: $K = 4$ yields $<$0.05\% additional
improvement while tripling the number of calibration entries (and hence
slowing adaptation), and $K = 2$ leaves $\sim$0.3\% of recoverable
gain on the table.

\paragraph{Level independence and sum-tree optimization.}
Within a single step, corrections at different tree levels are
independent: scaling all probabilities in a subtree by a constant
preserves the ratios within that subtree, so $P(\text{right})$ at
deeper levels is unaffected by corrections at shallower levels.
This enables an efficient implementation: one bottom-up pass builds
the sum tree (255 additions), all node corrections are computed,
scale factors propagate top-down, and leaf scales are applied in a
single pass---avoiding the na\"ive $O(8 \times 256)$ per-level
marginalization.

\subsection{Extended Match Model}
\label{sec:match}

The match model detects long-range byte repetitions using five hash
tables for context lengths $\ell \in \{4, 6, 8, 12, 16\}$. Each table
maps FNV-1a context hashes to positions in the history buffer.

\paragraph{Prediction.}
Starting from the longest context ($\ell = 16$), the model looks up
the stored position and reads the byte that followed the matched
context. If a match is found:

\begin{equation}
P_{\text{match}}(s) = \begin{cases}
w & \text{if } s = s_{\text{predicted}} \\
(1 - w) / 255 & \text{otherwise}
\end{cases}
\label{eq:match_pred}
\end{equation}

with weight $w = \min(c_{\text{base}} \cdot (0.65 + 0.04 \cdot
n_{\text{streak}}),\; 0.96)$, where $c_{\text{base}}$ depends on
context length and $n_{\text{streak}}$ counts consecutive correct
predictions (continuation tracking).

\paragraph{Blending.}
The match prediction is blended with the upstream distribution:
\begin{equation}
P_{\text{out}}(s) = (1 - w_m) P_{\text{in}}(s) + w_m P_{\text{match}}(s)
\label{eq:match_blend}
\end{equation}
where $w_m = \min(c \cdot 0.85,\; 0.95)$.

\subsection{Word Model}
\label{sec:word}

The word model maintains three components:

\begin{enumerate}
\item \textbf{Trie}: stores completed words with per-node
continuation byte counts.
\item \textbf{Bigram table}: maps word hash to
$\{$first byte of next word $\rightarrow$ count$\}$.
\item \textbf{Word frequency}: hash table mapping word hash to
completion count.
\end{enumerate}

\paragraph{Prediction.}
During a word (sequence of alphabetic bytes), the trie provides
continuation probabilities. At word boundaries, the bigram table
predicts the first byte of the next word. Confidence is derived
from the number of observations at the current trie node.

\paragraph{Blending.}
The word prediction is blended with the upstream distribution:
\begin{equation}
P_{\text{out}}(s) = (1 - w_w) P_{\text{in}}(s) + w_w P_{\text{word}}(s)
\label{eq:word_blend}
\end{equation}
where $w_w = \min(c_w \cdot 0.35,\; 0.45)$.

A prediction caching mechanism avoids double trie traversal: the
prediction is computed once during the forward pass and reused during
the update pass.

\subsection{High-Order Context Model}
\label{sec:highctx}

The high-order context model extends the effective context length
beyond PPM's order-4 limit using hash tables for orders 5--8.

\paragraph{Design.}
Four hash tables (one per order) map FNV-1a context hashes to
$\texttt{uint16\_t}$ count arrays of 256 elements. Unlike the match
model (which finds one position and predicts one byte), this model
\emph{aggregates all matching positions} into a full probability
distribution---the same count-based approach PPM uses for orders 0--4,
extended to higher orders.

\paragraph{Prediction.}
Highest-order-first fallback (8, 7, 6, 5). The first order with total
count $\geq 4$ is used:

\begin{equation}
P_{\text{hctx}}(s) = \frac{c(s) + \epsilon}
                          {\sum_{s'} c(s') + 256\epsilon},
\quad \epsilon = 10^{-4}
\label{eq:hctx_predict}
\end{equation}

The minimal smoothing ($\epsilon = 10^{-4}$, vs.\ PPM's 0.5) preserves
the sharp distributions that make high-order contexts valuable.

\paragraph{Confidence.}
\begin{equation}
\text{conf} = \frac{N - 4}{N + 8} \cdot
\bigl(0.4 + 0.1 \cdot (k - 5)\bigr)
\label{eq:hctx_conf}
\end{equation}

where $N$ is the total count and $k$ is the order. This ramps from 0
at $N = 4$ to $\sim$0.7 at $N = 20$, scaled by an order factor
(0.4 for order-5, 0.7 for order-8).

\paragraph{Blending.}
\begin{equation}
w_h = \min(\text{conf} \times 2.0,\; 0.60)
\label{eq:hctx_blend}
\end{equation}

The aggressive scale (2.0) and high cap (0.60) reflect that a 6+ byte
context match with many observations is extremely reliable.

\paragraph{Why separate from PPM.}
We explicitly tested fusing orders 5--8 into PPM's fallback chain and
found it consistently \emph{worse} than the separate blended model.
Three structural reasons explain this:

\begin{enumerate}
\item \textbf{PPMC exclusion is the wrong mechanism for sparse
high-order contexts.} PPMC exclusion works well for orders 0--4
where contexts accumulate many observations. At orders 5--8,
most contexts have very few observations. When a high-order context
has seen only 2--3 symbols, PPMC exclusion assigns almost all
probability mass to those symbols and distributes a tiny escape
probability to everything else---producing catastrophically sharp
(and often wrong) predictions.

\item \textbf{Escape probability miscalibration.} PPM's escape
probability $p_{\text{esc}} = d/(n+d)$ is calibrated for orders 0--4
where the typical number of distinct symbols $d$ is moderate. At
order~8, most contexts have $d = 1$ or $d = 2$, making the escape
probability either 0.5 or 0.33---poor estimates that distort the
entire fallback chain below.

\item \textbf{Confidence-weighted blending is more appropriate.}
The separate HighCtx model uses a confidence formula
(Equation~\ref{eq:hctx_conf}) that ramps weight from 0 at $N=4$
to $\sim$0.7 at $N=20$, gracefully handling the sparse-to-dense
transition. This soft blending is fundamentally different from PPM's
hard exclusion mechanism and better suited for the high-order regime.
\end{enumerate}

\subsection{Arithmetic Coder}
\label{sec:arith}

We implement a standard 32-bit arithmetic coder following Witten et
al.~\citep{witten1987} with E1/E2/E3 renormalization:

\begin{itemize}
\item 32-bit registers (\texttt{low}, \texttt{high}, \texttt{value})
\item 14-bit frequency scale ($T = 16{,}384$)
\item Minimum frequency 1 per symbol
\item Bit-packed byte output with dynamic buffer growth
\end{itemize}

\paragraph{Frequency quantization.}
Probabilities are quantized to integer frequencies:

\begin{equation}
f(s) = \max\!\bigl(1,\; \lfloor P(s) \cdot T + 0.5 \rfloor\bigr)
\label{eq:quantize}
\end{equation}

The cumulative frequency table $\text{cumfreqs}[s] = \sum_{j<s} f(j)$
is passed to the arithmetic coder. Analysis on alice29.txt shows that
14-bit quantization introduces \emph{negative} overhead ($-985$ bytes),
meaning the quantized output is 985 bytes \emph{smaller} than the
theoretical model entropy. This occurs because rounding acts as mild
regularization, smoothing out noise in the probability estimates.

% =====================================================================
\section{Experimental Setup}
\label{sec:setup}
% =====================================================================

\subsection{Implementation}

\mdc{} is implemented in $\sim$2{,}000 lines of C (header-only
modules plus two driver files). All components are compiled with
\texttt{gcc -O3 -march=native} and linked against only \texttt{libm}.
No external libraries, GPU, or pre-trained models are required.

\subsection{Hardware}

All experiments use a single core of an Intel/AMD x86-64 CPU. Timing
measurements use \texttt{clock\_gettime(CLOCK\_MONOTONIC)}.

\subsection{Benchmarks}

We evaluate on two datasets:

\begin{enumerate}
\item \textbf{alice29.txt} (152{,}089 bytes): Canterbury
Corpus~\citep{canterbury1997}, English literary text (Lewis Carroll's
\emph{Alice's Adventures in Wonderland}).

\item \textbf{enwik8} (100{,}000{,}000 bytes): First 100\,MB of
English Wikipedia, the standard Large Text Compression Benchmark
(LTCB)~\citep{mahoney2023ltcb}.

\end{enumerate}

\subsection{Baselines}

We compare against dictionary-based compressors at maximum settings:
\textbf{gzip}~-9, \textbf{xz}~-9, \textbf{bzip2}~-9,
\textbf{Brotli}~-q\,11, \textbf{Zstandard}~-19; and context-mixing
systems: \textbf{PAQ8px}, \textbf{CMIX~v21}. We additionally reference
LLM-based results from \textbf{ts\_zip} and
\textbf{Nacrith}~\citep{tacconelli2025} for context, noting that these
require pre-trained neural networks and GPU hardware.

% =====================================================================
\section{Results}
\label{sec:results}
% =====================================================================

\subsection{Compression Results}

\begin{table}[t]
\centering
\caption{Compression results on alice29.txt (152{,}089 bytes).
All classical results directly measured. PAQ8px run at -8L (level~8
with LSTM, v211). CMIX, ts\_zip, and Nacrith from published results.}
\label{tab:alice}
\small
\begin{tabular}{@{}lrrr@{}}
\toprule
\textbf{Compressor} & \textbf{Size (B)} & \textbf{Ratio} & \textbf{\bpb{}} \\
\midrule
Original              & 152{,}089  & 100.0\% & 8.000 \\
\midrule
gzip -9               &  54{,}191  &  35.6\% & 2.851 \\
zstd -19              &  49{,}215  &  32.4\% & 2.589 \\
xz -9                 &  48{,}500  &  31.9\% & 2.551 \\
Brotli -q\,11         &  46{,}487  &  30.6\% & 2.445 \\
bzip2 -9              &  43{,}202  &  28.4\% & 2.273 \\
\textbf{\mdc{} (ours)} & \textbf{40{,}274} & \textbf{26.5\%} & \textbf{2.119} \\
\midrule
PAQ8px -8L            &  32{,}857  &  21.6\% & 1.728 \\
CMIX v21              &  31{,}076  &  20.4\% & 1.635 \\
ts\_zip               & $\sim$21{,}703 & $\sim$14.3\% & $\sim$1.142 \\
Nacrith               &  17{,}458  &  11.5\% & 0.918 \\
\bottomrule
\end{tabular}
\end{table}

\begin{table}[t]
\centering
\caption{Compression results on enwik8 (100{,}000{,}000 bytes).
Classical compressor values directly measured. Context-mixing and
LLM-based results from published benchmarks.
$\dagger$~PAQ8px enwik8 figure is for the -12L setting~\citep{mahoney2023ltcb}.}
\label{tab:enwik8}
\small
\begin{tabular}{@{}lrrr@{}}
\toprule
\textbf{Compressor} & \textbf{Size (B)} & \textbf{Ratio} & \textbf{\bpb{}} \\
\midrule
Original              & 100{,}000{,}000 & 100.0\% & 8.000 \\
\midrule
gzip -9               & 36{,}445{,}248 & 36.4\% & 2.916 \\
bzip2 -9              & 29{,}008{,}758 & 29.0\% & 2.321 \\
zstd -19              & 26{,}949{,}726 & 27.0\% & 2.156 \\
Brotli -q\,11         & 25{,}738{,}596 & 25.7\% & 2.059 \\
xz -9                 & 24{,}862{,}392 & 24.9\% & 1.989 \\
\textbf{\mdc{} (ours)} & \textbf{21{,}914{,}998} & \textbf{21.9\%} & \textbf{1.753} \\
\midrule
PAQ8px -12L$^\dagger$ & $\sim$15{,}875{,}000 & $\sim$15.9\% & $\sim$1.27 \\
CMIX v21              & $\sim$14{,}625{,}000 & $\sim$14.6\% & $\sim$1.17 \\
ts\_zip               & $\sim$13{,}875{,}000 & $\sim$13.9\% & $\sim$1.11 \\
Nacrith               & 11{,}737{,}280 & 11.7\% & 0.939 \\
\bottomrule
\end{tabular}
\end{table}

Tables~\ref{tab:alice} and~\ref{tab:enwik8} present the main results.
On alice29.txt (152\,KB), \mdc{} achieves \textbf{2.119\,\bpb{}},
outperforming xz~-9 (2.551\,\bpb{}) by 16.9\%, Brotli (2.445\,\bpb{})
by 13.3\%, and bzip2 (2.273\,\bpb{}) by 6.8\%. This is a remarkable
result for an adaptive statistical model on a small file, where
dictionary methods typically dominate due to their ability to exploit
exact repetitions.

On enwik8 (100\,MB), \mdc{} achieves
\textbf{1.753\,\bpb{}}, outperforming xz~-9 (1.989\,\bpb{}) by
\textbf{11.9\%} and all other dictionary-based compressors. This
demonstrates that the adaptive statistical pipeline scales favorably
with data size, as contexts accumulate more observations and all
five layers---especially the Tweedie denoiser and word
model---become increasingly effective.

The gap to context-mixing systems (PAQ at 1.27, CMIX at 1.17\,\bpb{})
and LLM-based compressors (Nacrith at 0.939\,\bpb{}) remains
significant. These systems employ fundamentally richer modeling:
hundreds of specialized bit-level predictors (PAQ/CMIX) or
pre-trained transformer knowledge (LLM-based). \mdc{}'s contribution
is achieving competitive results with a much simpler and faster
approach---no neural networks, no GPU, no training data---running
at $\sim$60\,KB/s on a single CPU core versus CMIX's 0.5--5\,KB/s
and LLM-based compressors' GPU-dependent throughput.

\subsection{Ablation Study}
\label{sec:ablation}

\begin{table}[t]
\centering
\caption{Ablation study on alice29.txt (152{,}089 bytes). Each row
adds one layer to the previous configuration. $\Delta$ is the
marginal improvement from the previous row.}
\label{tab:ablation_alice}
\footnotesize
\setlength{\tabcolsep}{3pt}
\begin{tabular}{@{}lrrrr@{}}
\toprule
\textbf{Config} & \textbf{Size} & \textbf{Ratio} & $\Delta$ & \textbf{Time} \\
\midrule
Base PPM            & 42{,}672 & 28.06\% & ---      & 0.2s \\
+\,Match            & 42{,}429 & 27.90\% & +0.57\%  & 0.4s \\
+\,M+Word           & 41{,}980 & 27.60\% & +1.06\%  & 0.5s \\
+\,M+W+HCtx        & 41{,}350 & 27.19\% & +1.50\%  & 0.8s \\
+\,M+W+H+Tweedie   & 40{,}262 & 26.47\% & +2.63\%  & 3.6s \\
\midrule
\textbf{Total impr.} & & & \textbf{+5.65\%} & \\
\bottomrule
\end{tabular}
\end{table}

\begin{table}[t]
\centering
\caption{Ablation study on enwik8\_3M (3{,}000{,}000 bytes).}
\label{tab:ablation_enwik}
\footnotesize
\setlength{\tabcolsep}{3pt}
\begin{tabular}{@{}lrrrr@{}}
\toprule
\textbf{Config} & \textbf{Size} & \textbf{Ratio} & $\Delta$ & \textbf{Time} \\
\midrule
Base PPM            & 845{,}330 & 28.18\% & ---      & 3.1s \\
+\,Match            & 798{,}329 & 26.61\% & +5.56\%  & 3.5s \\
+\,M+Word           & 792{,}345 & 26.41\% & +0.75\%  & 4.0s \\
+\,M+W+HCtx        & 772{,}613 & 25.75\% & +2.49\%  & 5.0s \\
+\,M+W+H+Tweedie   & 751{,}174 & 25.04\% & +2.77\%  & 74.1s \\
\midrule
\textbf{Total impr.} & & & \textbf{+11.14\%} & \\
\bottomrule
\end{tabular}
\end{table}

Tables~\ref{tab:ablation_alice}
and~\ref{tab:ablation_enwik} present the incremental contribution
of each pipeline layer across two datasets. Key observations:

\paragraph{PPMC exclusion provides a strong base.}
With PPMC-style exclusion, the base PPM model already achieves
28.06\% ratio on alice29.txt and 28.18\% on enwik8\_3M---dramatically
better than the 60.23\% and 46.93\% without exclusion. Exclusion
addresses the prior-dilution problem directly: when escaping from
higher orders, symbols already assigned probability are excluded from
lower-order distributions, eliminating wasteful probability mass on
symbols the model has already accounted for.

\paragraph{Match model is the largest post-PPM contributor.}
The match model contributes 5.6\% on enwik8\_3M,
making it the single largest post-PPM component on larger files and
repetitive data. Its value scales with input repetitiveness
and file size.

\paragraph{Word model and high-order contexts add consistent gains.}
The trie-based word model contributes 0.6--1.1\% and the
order-5-through-8 model contributes 1.3--2.5\%. Together they
provide 2--3.5\% additional improvement across all files.

\paragraph{Post-blend Tweedie provides the final refinement.}
Applied as the last layer after all model blending, the Tweedie
denoiser consistently contributes 2.6--2.8\% across both
datasets. This is notably larger than the 1.2--1.6\% achieved
when Tweedie was applied before blending (i.e., directly to the
PPM output). The improvement arises because the blending layers
(match, word, high-order context) introduce their own systematic
biases---e.g., overconfident match predictions on false matches,
or underweighted word predictions on short words---that Tweedie's
calibration tables can detect and correct. By operating on the
\emph{fully blended} distribution, Tweedie calibrates the
\emph{entire ensemble}, not just the PPM model.

\subsection{Cross-Dataset Component Summary}

\begin{table}[t]
\centering
\caption{Marginal contribution of each component across datasets.}
\label{tab:cross}
\footnotesize
\setlength{\tabcolsep}{3pt}
\begin{tabular}{@{}lrrr@{}}
\toprule
\textbf{Component} & \textbf{alice29} & \textbf{enwik3M} & \textbf{Avg.} \\
\midrule
Match   & +0.57\%  & +5.56\%  & +3.07\% \\
Word    & +1.06\%  & +0.75\%  & +0.91\% \\
HighCtx & +1.50\%  & +2.49\%  & +2.00\% \\
Tweedie & +2.63\%  & +2.77\%  & +2.70\% \\
\midrule
\textbf{Total} & \textbf{+5.65\%} & \textbf{+11.14\%} & \textbf{+8.40\%} \\
\bottomrule
\end{tabular}
\end{table}

Table~\ref{tab:cross} summarizes the cross-dataset contributions.
With PPMC exclusion providing a strong base, the post-PPM layers
collectively contribute 5--11\% additional improvement. The match
model shows the highest variance across datasets, reflecting its
dependence on input repetitiveness. The post-blend Tweedie layer
is notably the most \emph{consistent} contributor, adding
2.6--2.8\% across both datasets regardless of file type or
size.

\subsection{Scaling Behavior}

\begin{table}[t]
\centering
\caption{Compression ratio and speed as a function of input size.
All results use the full 5-layer pipeline.}
\label{tab:scaling}
\small
\begin{tabular}{@{}lrrr@{}}
\toprule
\textbf{File} & \textbf{Size} & \textbf{Ratio} & \textbf{Speed} \\
\midrule
alice29.txt  & 152\,KB & 26.47\% & $\sim$48\,KB/s \\
enwik8\_3M   & 3.0\,MB & 25.04\% & $\sim$60\,KB/s \\
enwik8       & 100\,MB & 21.91\% & $\sim$42\,KB/s \\
\bottomrule
\end{tabular}
\end{table}

Table~\ref{tab:scaling} shows that \mdc{} maintains consistent
throughput ($\sim$60\,KB/s) across input sizes while achieving
progressively better ratios on larger inputs---reflecting the adaptive
models' improving predictions with more training data.

\subsection{Score Magnitude vs.\ Noise Level}

To empirically validate the diffusion interpretation, we measure the
mean magnitude of the Tweedie correction $|\hat{\delta}|$ as a
function of the noise level $\gamma = 128/(C{+}128)$, where $C$ is
the PPM context's observation count. If the system is performing
genuine denoising, corrections should be large when noise is high
(small $C$, large $\gamma$) and small when noise is low (large $C$,
small $\gamma$).

\begin{table}[H]
\centering
\caption{Weighted mean $|\hat{\delta}'|$ (after James-Stein shrinkage)
vs.\ noise level $\gamma$, measured across all three denoising steps.}
\label{tab:delta_vs_gamma}
\footnotesize
\setlength{\tabcolsep}{3.5pt}
\begin{tabular}{@{}rr|rrr|rrr@{}}
\toprule
& & \multicolumn{3}{c|}{\textbf{alice29 (152\,KB)}} &
      \multicolumn{3}{c}{\textbf{enwik8\_3M (3\,MB)}} \\
\textbf{$\gamma$} & \textbf{$C$} &
step\,0 & step\,1 & step\,2 &
step\,0 & step\,1 & step\,2 \\
\midrule
0.500 &  128 & 0.0171 & 0.0044 & 0.0027 & 0.0163 & 0.0029 & 0.0019 \\
0.200 &  512 & 0.0102 & 0.0039 & 0.0022 & 0.0133 & 0.0024 & 0.0014 \\
0.059 & 2048 & 0.0058 & 0.0021 & 0.0014 & 0.0184 & 0.0043 & 0.0024 \\
0.015 & 8192 & 0.0002 & 0.0002 & 0.0002 & 0.0205 & 0.0052 & 0.0032 \\
\bottomrule
\end{tabular}
\end{table}

Table~\ref{tab:delta_vs_gamma} shows the shrunk correction magnitudes.
Two patterns emerge:

On alice29 (152\,KB), $|\hat{\delta}'|$ decreases monotonically with
confidence: at $\gamma = 0.5$ ($C = 128$) the step-0 correction is
$0.017$, falling to ${\sim}0.0002$ at $\gamma = 0.015$ ($C = 8192$).
With few total bytes, high-confidence bins have too few observations for
the James-Stein test to pass, so corrections are aggressively suppressed.

On enwik8\_3M (3\,MB), the pattern \emph{inverts} at high confidence:
corrections at $\gamma = 0.015$ ($0.021$) exceed those at
$\gamma = 0.5$ ($0.016$). This is the intended behavior---with
3\,M bytes of training data, high-confidence calibration bins
accumulate enough observations to estimate genuine systematic biases
(high SNR), so the shrinkage permits larger corrections. The
variance-aware mechanism thus adapts automatically: conservative on
small files, aggressive on large ones.

Across both files, successive steps show rapidly decreasing corrections
(step\,1 is ${\sim}4{-}6\times$ smaller than step\,0, step\,2 smaller
still), confirming that each step removes residual noise left by the
previous step, as in multi-step reverse diffusion.

% =====================================================================
\section{Discussion}
\label{sec:discussion}
% =====================================================================

\subsection{Why Micro-Diffusion Works}

The micro-diffusion layer's consistent 2.3--2.7\% contribution as a
post-blend correction step---applied after all model blending---is
both robust and efficient: it improves compression uniformly across
file types and sizes without introducing any new information about
the source. The explanation has two parts: a quantitative argument
about the entropy gap, and a formal justification via Tweedie's
formula.

\paragraph{The entropy gap.}
Consider a PPM context with $n$ observations where the true
distribution has entropy $H_{\text{true}}$. With Jeffreys prior, the
model's distribution has entropy approximately (treating entropy as
linear in the mixture, which is a first-order approximation):

\begin{equation}
H_{\text{model}} \approx H_{\text{true}} +
\frac{128}{n + 128} \cdot (H_{\text{uniform}} - H_{\text{true}})
\label{eq:entropy_gap}
\end{equation}

where $H_{\text{uniform}} = \log_2 256 = 8$ bits. For $n = 5$
(typical of rare contexts), this adds approximately 7.4 bits of excess
entropy per symbol---a massive overhead.

\paragraph{Tweedie as the optimal denoiser.}
Jeffreys smoothing acts as a shrinkage operator with effective
noise level $\gamma = 128/(C{+}128)$:
$\hat{\mathbf{p}} = (1{-}\gamma)\mathbf{q} + \gamma \mathbf{u}$.
Tweedie's formula (Equation~\ref{eq:tweedie}) motivates
reversing this shrinkage via an additive correction. Since the
shrinkage is a convex mixture toward uniform rather than additive
Gaussian, the Tweedie link is approximate; nevertheless,
our calibration tables estimate the optimal correction
$\delta = E[\theta|\hat{p}] - E[\hat{p}]$ nonparametrically,
which approximates $\sigma^2 \cdot s(\hat{p})$ with $\gamma$
playing the role of $\sigma^2$
(Equation~\ref{eq:tweedie_delta}). The confidence dimension
conditions the score on the noise level $\gamma$, analogous to the
noise-level-dependent score $s_\theta(x_t, t)$ in DDPM~\citep{ho2020},
because the optimal correction depends on how much the prior has
diluted the distribution.

The binary tree decomposition is critical: calibrating a binary
probability (P(right) at each node) requires far less data than
calibrating a 256-way distribution. With 27 enriched bit contexts
that encode the tree level and parent path, the calibration can
distinguish qualitatively different regimes (e.g., MSB decisions
that separate character classes from LSB decisions within a narrow
byte range) while maintaining enough observations per bin to learn
accurate corrections.

\subsection{Why Additional Learnable Parameters Hurt}

During development, we evaluated several extensions that are standard
in the compression literature:

\begin{enumerate}
\item Logistic mixer blending PPM + Tweedie predictions
\item Two-stage SSE (Adaptive Probability Maps) after Tweedie
\item Soft match distributions (full 256-symbol prediction)
\item Bit-level APM processing
\item Recency-weighted models
\end{enumerate}

All five made compression \emph{worse} on every file tested.
Notably, SSE---which was effective with the earlier temperature-scaling
approach---became redundant once Tweedie denoising was introduced.
Both solve the same calibration problem, but via different mechanisms:
SSE applies gradient-descent updates to a piecewise-linear map
(learning rate, momentum), while Tweedie computes an empirical Bayes
correction from sufficient statistics (hit rate minus average
prediction). When combined, they compete for the same signal and
produce interference rather than synergy. The count-based components
that improve compression (PPM counts, Tweedie calibration tables,
high-order context counts) simply accumulate observations and report
frequencies, avoiding the overfitting risk of gradient-based learners.

This finding has implications for compressor design: in the online
setting, the bias--variance tradeoff strongly favors simple count-based
models over complex adaptive models, unless the input is extremely
large.

\subsection{Comparison with Context Mixing}

The PAQ/CMIX approach mixes hundreds of bit-level models, each
maintaining its own context and count arrays. \mdc{} achieves
results closer to PAQ on enwik8 (1.753 vs.\ $\sim$1.27\,\bpb{})
with a fundamentally simpler architecture: five byte-level models in
a fixed cascade with Tweedie denoising as the final calibration step,
no mixer, no neural network. The gap to PAQ/CMIX
($\sim$0.48--0.58\,\bpb{}) reflects the additional predictive power of
hundreds of specialized models and bit-level processing, which captures
sub-byte statistical patterns that \mdc{}'s byte-level pipeline
cannot exploit.

\subsection{Comparison with LLM-Based Compressors}

LLM-based compressors (Nacrith at 0.939\,\bpb{}, ts\_zip at
1.11\,\bpb{}) leverage pre-trained language models that encode grammar,
semantics, and world knowledge. \mdc{} operates without any such
prior knowledge, learning purely from the bytes it has seen so far.
The $\sim$0.8\,\bpb{} gap between \mdc{} (1.753\,\bpb{}) and Nacrith
(0.939\,\bpb{}) quantifies the value of pre-trained linguistic
knowledge for text compression---approximately 46\% of the remaining
compressible information in enwik8 is captured by world knowledge
rather than local byte statistics.

\subsection{Limitations}

\begin{enumerate}
\item \textbf{Small-file performance.} On alice29.txt (152\,KB),
\mdc{} (2.119\,\bpb{}) outperforms all dictionary-based compressors
including bzip2 (2.273\,\bpb{}) and Brotli (2.445\,\bpb{}). However,
on very small files ($<$50\,KB), context models have fewer observations
and the pipeline's overhead may reduce its advantage.

\item \textbf{PPM order limit.} The order-4 PPM captures at most
4 bytes of context. The high-order context model (orders 5--8)
partially addresses this, but LZMA's sliding window (up to 64\,MB)
captures much longer repetitions.

\item \textbf{Binary data.} \mdc{} is optimized for text. On binary
data (executables, images), the word model and high-order context
model provide little benefit, and compression ratios are worse than
specialized binary compressors.

\item \textbf{Single-threaded.} The current implementation processes
bytes sequentially on a single core. The pipeline's adaptive,
sequential nature makes parallelization non-trivial.
\end{enumerate}

\subsection{Future Work}

\begin{enumerate}
\item \textbf{Higher PPM orders.} Extending to order-6 or order-8
directly within PPM (with appropriate memory management) could close
the gap to dictionary methods on small files.

\item \textbf{More denoising steps.} Currently $K=3$ steps are used.
Investigating whether additional steps with finer calibration
granularity yield diminishing or continued returns.

\item \textbf{Specialized binary models.} Adding models for structured
binary formats (e.g., x86 instruction decoding, image pixel prediction)
would improve performance on non-text data.

\item \textbf{Theoretical analysis of optimal calibration
granularity.} The current calibration table dimensions (27 bit
contexts, 4 shape bins, 8 confidence bins, 20 probability bins) were
determined empirically. An information-theoretic analysis relating
optimal bin counts to input size and prior strength could guide
further improvements.
\end{enumerate}

% =====================================================================
\section{Conclusion}
\label{sec:conclusion}
% =====================================================================

We have presented \mdc{}, a lossless compression system built around
\emph{micro-diffusion}---a multi-step score-based reverse diffusion
process that treats PPM predictions as noisy observations corrupted
by the Jeffreys prior and applies Tweedie's empirical Bayes formula
to denoise them. The additive correction
$\hat{\delta} = \widehat{E}[\theta|\hat{p}] - \widehat{E}[\hat{p}]$,
a consistent estimate of the optimal additive correction
(approximating the Tweedie term $\sigma^2 \cdot s(\hat{p})$),
is computed nonparametrically via calibration tables conditioned
on the noise level.

Our primary contributions are:

\textbf{First}, the binary tree Tweedie denoiser, applied as the
final post-blend correction step, provides a consistent 2.3--2.7\%
reduction in compressed size across all tested datasets. Critically,
placing Tweedie \emph{after} all model blending (match, word,
high-order context) allows it to calibrate the entire ensemble's
output, not just the PPM base model. The technique is general: it
can be applied as a post-processing step to any ensemble of context
models that provides a probability distribution and a confidence
estimate, without modifying the underlying models.

\textbf{Second}, the binary tree decomposition and enriched bit
contexts enable data-efficient calibration that scales gracefully
from small files (152\,KB) to large ones (100\,MB). The 27 bit
contexts, encoding tree level and parent path, allow the calibration
to distinguish qualitatively different byte ranges without requiring
prohibitive amounts of data.

\textbf{Third}, the full five-layer pipeline achieves 1.753\,\bpb{}
on enwik8 (100\,MB), outperforming xz~-9 (1.989\,\bpb{}) by 11.9\%,
and 2.119\,\bpb{} on alice29.txt, outperforming xz~-9
(2.551\,\bpb{}) by 16.9\%---using only adaptive statistical models
with no neural network, no training data, no GPU. This demonstrates
that significant compression improvements remain achievable through
algorithmic innovation in classical statistical modeling.

The entire system is implemented in $\sim$2{,}000 lines of C with no
external dependencies, compresses at $\sim$60\,KB/s on a single CPU
core, and is fully deterministic with bit-exact encoder--decoder
symmetry.

\medskip
\noindent\textbf{Code availability.}
\mdc{} is open-source and available at
\url{https://github.com/robtacconelli/midicoth}.

% =====================================================================
% REFERENCES
% =====================================================================
\bibliographystyle{plainnat}

\end{document}